\documentclass[sigconf]{acmart}

\AtBeginDocument{%
  }

\usepackage{booktabs}
\usepackage[table]{xcolor}
\usepackage{graphicx}
\usepackage{array}

\usepackage{makecell}

\copyrightyear{2026}
\acmYear{2026}
\setcopyright{cc}
\setcctype{by}
\acmConference[CIKM '26]{Proceedings of the 35th ACM International Conference on Information and Knowledge Management}{November 07--11, 2026}{Rome, Italy}
\acmBooktitle{Proceedings of the 35th ACM International Conference on Information and Knowledge Management (CIKM '26), November 07--11, 2026, Rome, Italy}
\acmDOI{10.1145/3799682.3841123}
\acmISBN{979-8-4007-2539-5/2026/11}

\begin{document}

\title{RI-Prune: Relation and Information Aware Visual Token Pruning for Multi-Image Vision-Language Models}

\author{Dongjin Lee}
\orcid{0009-0002-6802-5186}
\affiliation{%
  \institution{Chung-Ang University}
  \city{Seoul}
  \country{Republic of Korea}
}
\email{dongjinlee30@cau.ac.kr}

\author{Eunju Lee}
\orcid{0000-0002-6571-0156}
\affiliation{%
  \institution{Chung-Ang University}
  \city{Seoul}
  \country{Republic of Korea}
}
\email{dmswn5829@cau.ac.kr}

\author{Jinhee Jang}
\orcid{0009-0007-4194-7099}
\affiliation{%
  \institution{Chung-Ang University}
  \city{Seoul}
  \country{Republic of Korea}
}
\email{jinheejang@cau.ac.kr}

\author{Yoonji Lee}
\orcid{0009-0006-9108-3252}
\affiliation{%
  \institution{Chung-Ang University}
  \city{Seoul}
  \country{Republic of Korea}
}
\email{pioneer0305@cau.ac.kr}

\author{Youngbin Kim}
\orcid{0000-0002-2114-0120}
\authornote{Corresponding author}
\authornotemark[0]
\affiliation{%
  \institution{Chung-Ang University}
  \city{Seoul}
  \country{Republic of Korea}
}
\email{ybkim85@cau.ac.kr}



\begin{abstract}
    Large vision-language models (LVLMs) have recently been extended to handle multiple input images, but this extension dramatically increases the visual token budget, making visual-token pruning not merely an efficiency option but a practical necessity. Existing visual-token pruning methods, primarily designed for the single-image setting, rely on attention-derived saliency signals to estimate per-token importance. However, this attention-centric design transfers poorly to the multi-image regime: it causes substantial performance degradation on multi-image benchmarks, suffers from strong positional bias, and loses discriminative attention signals as the number of input images increases. To test whether attention-selected tokens are truly important for multi-image reasoning, we conduct two complementary causal probes—Leave-One-In and Leave-One-Deletion—which measure token importance by restoring individual tokens to a zeroed visual sequence or removing them from the full sequence. Attention-based pruners provide only marginal gains over random selection, showing that attention is a weak proxy for visual-token importance in multi-image reasoning. Motivated by this gap, we propose Relation and Information Aware Pruning (RI-Prune), a training-free, attention-free pre-decoder selector for multi-image visual token pruning. RI-Prune scores each visual token using a z-normalized combination of three signals computed from projector outputs: query relevance, cross-image correspondence with compact per-image prototypes, and intra-image representativeness. By integrating these signals, RI-Prune performs balanced per-image Top-$K$ selection to preserve information across input images. Across multiple multi-image benchmarks and three LVLM backbones, RI-Prune achieves a stronger trade-off between token reduction and task performance than prior pruning methods, particularly under aggressive token budgets.
\end{abstract}



\maketitle

\section{Introduction}
The rapid evolution of Large Vision-Language Models (LVLMs)~\cite{li2024llava,bai2023qwen,yang2025qwen3, li2024llavaonevision} has enabled a significant transition from single-image processing to sophisticated multi-image reasoning~\cite{awadalla2023openflamingo,jiang2024mantis,laurencon2024idefics}. Modern architectures are now capable of performing diverse tasks such as identifying differences or commonalities across multiple visual inputs and understanding temporal sequences~\cite{maaz2024videochatgpt,lin2024videollava}. These advancements allow models to integrate information from various sources to provide comprehensive multimodal responses. However, the increasing number of images leads to a proportional surge in the volume of vision tokens which creates substantial challenges regarding computational cost and inference latency~\cite{li2024llava,kim2024token,bai2023qwen,zhu2024vision}. Since the self-attention mechanism in the Transformer backbone exhibits quadratic complexity~\cite{keles2023computational,shah2024flashattention} relative to the sequence length, the inclusion of multiple high-resolution images often results in an unsustainable burden on hardware resources.

\begin{figure*}[t]
    \centering
    \includegraphics[width=\textwidth]{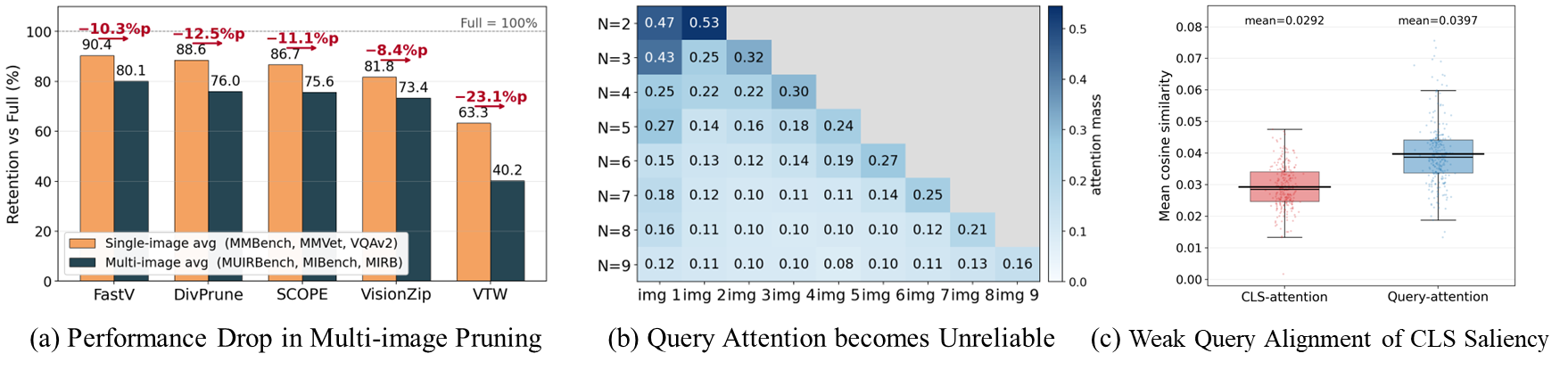}
    \caption{
    Failure modes of attention-based visual token pruning in multi-image LVLMs.
    \textbf{(a)} Single-image pruning methods retain high performance on single-image benchmarks but degrade sharply on multi-image benchmarks, showing poor transfer to multi-image reasoning.
    \textbf{(b)} Query attention becomes unreliable across images: it exhibits positional bias and becomes increasingly non-discriminative as the number of input images grows.
    \textbf{(c)} CLS-based saliency selects tokens with weaker query alignment than query-driven attention, suggesting that CLS attention is less effective at preserving question-relevant visual evidence.
    }
    \label{fig:fig1}
\end{figure*}

To address efficiency challenges, various pruning methods have been proposed to remove redundant visual information~\cite{qian2026swiftvlm, kim2024token, lin2025boosting, yin2022vit, liang2022not, ye2025fit, fan2025visipruner}. These methods primarily rely on attention-based importance scores to identify important visual tokens, which broadly fall into two categories: query-driven selection in the LLM decoder, where visual tokens are ranked by the attention they receive from instruction tokens at an intermediate layer~\cite{chen2024image, zhang2025sparsevlm, xing2025conical, deng2025scope, huang2024ivtp} and Class token (CLS)-driven selection in the vision encoder, where each patch is scored by its CLS token-to-patch attention~\cite{yang2025visionzip, shang2025llavaprumerge, zhang2025vispruner}. Although these attention-based signals are effective in the single-image setting, it remains unclear whether they can reliably identify important visual evidence when multiple images must be jointly reasoned over. 

To examine this, we analyze their behavior in multi-image LVLMs and identify three concrete failure modes, as shown in Figure~\ref{fig:fig1}. First, existing single-image pruners retain 82.2\% accuracy on average on single-image benchmarks~\cite{liu2024mmbench, yu2024mmvet, goyal2017making}, but drop to 69.1\% on multi-image benchmarks~\cite{wang2025muirbench, liu2024mibench, zhao2024mirb}, yielding a 13.1\%p degradation. This gap indicates that pruning strategies effective for single-image inputs do not transfer reliably to multi-image scenarios. Second, the instruction-to-image attention exploited by query-driven pruners exhibits a strong positional bias: boundary images receive disproportionately large attention, whereas images in the middle are underrepresented, resembling a lost-in-the-middle effect. As the number of images $N$ grows, the per-image attention mass further collapses toward a nearly uniform distribution. This
makes attention signals less reliable as indicators of token importance, especially in multi-image settings. Third, compared with query-driven attention, CLS-based saliency selects tokens that have lower cosine similarity to the query embedding. This suggests that CLS-based pruning is less sensitive to the question and is therefore less capable of preserving query-relevant evidence in multi-image reasoning. Consequently, attention magnitudes computed within a single image---whether from the LLM decoder or the vision encoder's CLS token---fail to provide a reliable basis for multi-image visual token pruning. 
This motivates the need for an attention-free pruning strategy that preserves vision tokens essential for both cross-image relationships and question-relevant evidence in multi-image reasoning.

To verify whether the tokens prioritized by existing pruners truly carry the
evidence needed for multi-image reasoning, we design two complementary causal
token-importance probes: \emph{Leave-One-In}, which measures each token's
standalone evidence by restoring it into a fully zeroed visual sequence, and
\emph{Leave-One-Deletion}, which measures its marginal contribution by removing
it from the full visual sequence. Aggregating these per-token information
values over each pruner's retained set, we find that attention-based
pruners--including FastV-style query attention~\cite{chen2024image} and VisionZip-style CLS
saliency~\cite{yang2025visionzip}--achieve only marginal gains over uniformly random selection under
the same budget. This indicates that intra-image attention magnitudes are an
unreliable proxy for token utility in multi-image reasoning, motivating a new
selection criterion beyond attention-based saliency.


Motivated by these findings, we propose \textbf{R}elation and  \textbf{I}nformation Aware Pruning (\textbf{RI-Prune}), a training-free,
attention-free visual token pruning method for multi-image LVLMs. RI-Prune
avoids unstable attention magnitudes by estimating token importance directly in
the projector embedding space before decoding. RI-Prune scores each visual
token using three complementary signals: query relevance, cross-image
correspondence, and intra-image representativeness. Query relevance captures
direct task conditioning by measuring the maximum cosine similarity between
each visual token and the question-token embeddings. Cross-image correspondence
identifies relational evidence by comparing each token with compact per-image prototypes from the other images. Intra-image representativeness
preserves image-specific evidence by favoring tokens that are both strongly
aligned with local prototypes and assigned to less frequent visual patterns,
thereby retaining distinctive yet coherent regions while reducing repeated
background cues. These signals are sample-wise $z$-normalized and fused into a
unified importance score, from which RI-Prune performs per-image Top-$K$
selection by evenly allocating the token budget across input images and
retaining the highest-scoring tokens within each image, ensuring every image
contributes visual evidence under aggressive budgets. Experiments on multiple
multi-image benchmarks with Qwen2.5-VL-7B~\cite{bai2025qwen25vl},
Qwen3-VL-8B~\cite{yang2025qwen3}, and
LLaVA-OneVision-7B~\cite{li2024llavaonevision} show that RI-Prune provides a
robust trade-off between token reduction and task performance. In particular,
it achieves favorable performance compared with prior pruning methods on
relation-heavy tasks and maintains strong performance retention under
aggressive token budgets. The main contributions of this work are summarized as follows:

\begin{itemize}
    \item We reveal the limitations of attention-based visual token pruning in multi-image LVLMs. We show that query-driven pruning suffers from positional bias and attention dilution as the number of images increases, while CLS-driven pruning exhibits weak query alignment. Both types of attention-based pruning fail to reliably identify visual tokens required for correct multi-image reasoning.

    \item We propose \textbf{R}elation and \textbf{I}nformation Aware Pruning (RI-Prune), a novel attention-free visual token pruning strategy for multi-image LVLMs. RI-Prune operates directly in the projector embedding space and jointly models query relevance, cross-image correspondence, and intra-image representativeness, which are fused through sample-wise $z$-score normalization and weighted fusion to preserve tokens that are both question-aligned and relationally informative.

    \item We integrate RI-Prune into representative multi-image LVLMs, including Qwen2.5-VL-7B~\cite{bai2025qwen25vl}, Qwen3-VL-8B~\cite{yang2025qwen3}, and LLaVA-OneVision-7B~\cite{li2024llavaonevision}, without any additional training. Extensive experiments on multiple multi-image vision-language benchmarks demonstrate that RI-Prune achieves a favorable trade-off between computational efficiency and task performance, particularly under aggressive token budgets.
\end{itemize}

\section{Related Work}

\noindent\textbf{Large Vision-Language Models.} Following the paradigm shift initiated by Mantis-Eval~\cite{jiang2024mantis}, which established a framework for multi-image reasoning through large-scale training on interleaved image–text data, recent LVLMs such as Qwen2.5-VL~\cite{bai2025qwen25vl} and LLaVA-OneVision~\cite{li2024llavaonevision} have further advanced these capabilities. These models introduce architectures designed to process interleaved visual–text sequences while maintaining spatial and temporal coherence across multiple images. For instance, Qwen2.5-VL employs a dynamic-resolution Vision Transformer with Multimodal Rotary Position Embedding (M-ROPE) to encode spatial layouts and temporal order. Although it adopts a $2 \times 2$ spatial merging strategy to reduce token density, the sequence length can still grow substantially with multiple high-resolution images. Similarly, LLaVA-OneVision uses the AnyRes strategy, which decomposes images into local tiles and a global view, leading to a rapid increase in vision tokens as the number of images grows. While these designs improve structural representation and scalability, they primarily emphasize spatial preservation rather than addressing semantic redundancy or cross-image relationships. Consequently, the quadratic complexity of self-attention over long sequences still imposes significant computational overhead in multi-image LVLMs~\cite{lin2024vila,liu2023ring}.

\noindent\textbf{Pruning Methods in LVLMs.} The increasing complexity of LVLMs has driven the development of efficient token reduction methods to alleviate computational overhead~\cite{shah2024flashattention, lieber2024jamba}. Early work such as ToME~\cite{bolya2022token} reduces sequence length by merging similar visual tokens. Later studies focus on pruning redundant tokens within a single image. FastV~\cite{chen2024image} removes low-importance tokens by aggregating the attention each token receives from non-visual tokens at a shallow decoder layer, while VisionZip~\cite{yang2025visionzip} instead performs encoder-side selection, scoring patches by the attention they receive in the last vision-encoder block before the language model. VTW~\cite{lin2025boosting} discards visual tokens entirely after a fixed decoding layer so that subsequent reasoning relies only on text tokens, and HiRED~\cite{arif2025hired} allocates token budgets hierarchically across sub-regions of high-resolution inputs via vision-encoder attention. To move beyond pure saliency, coverage-aware methods have also been explored: DivPrune~\cite{alvar2025divprune} selects a maximally diverse subset of visual tokens via greedy max--min sampling in the embedding space, and SCOPE~\cite{deng2025scope} jointly models saliency and coverage of the retained tokens to preserve their semantic completeness. More recent methods improve adaptivity: PACT~\cite{dhouib2025pact} estimates token importance without attention scores, enabling compatibility with efficient kernels such as FlashAttention~\cite{shah2024flashattention}, and SwiftVLM~\cite{qian2026swiftvlm} introduces cross-layer token bypass to reduce the risk of losing important information. However, these methods are largely designed for single-image settings and mainly target intra-image redundancy, limiting their ability to capture the cross-image relations required for multi-image reasoning.

\section{Causal Analysis of Visual Token Information}

\subsection{Preliminaries.}

Modern LVLMs such as Qwen2.5-VL~\cite{bai2025qwen25vl} and LLaVA-OneVision
~\cite{li2024llavaonevision} typically utilize a modular architecture comprising a vision encoder, a connector, and an LLM backbone. Visual features extracted by the encoder are mapped into vision tokens by the connector through MLP or resampling mechanisms to serve as soft prompts for the LLM. In multi-image scenarios, these models process multiple inputs by concatenating their respective vision token sequences in an interleaved or sequential format following the order of system prompts, interleaved images, and the user question~\cite{li2024llava}. This integrated multimodal input constitutes the sequence $S$. These models inherit the autoregressive nature of Transformer-based LLMs where the generation of a response sequence $Y = \{y_1, y_2, \dots, y_{L_y}\}$, representing a set of ${L_y}$ predicted tokens given the input sequence $S$, is factorized as follows:
\[
P(Y \mid S) = \prod_{i=1}^{L_y} P(y_i \mid S, y_{<i})
\]
Under this formulation, each individual token $y_i$ is generated conditionally based on the initial multimodal context and the set of all preceding output tokens denoted as $y_{<i}$. This modeling paradigm is strictly enforced by a causal attention mechanism that ensures each token can only attend to its preceding positions in the sequence.

\subsection{Analysis Protocol.}

To examine whether a pruning method preserves visual tokens that are truly
important for multi-image reasoning, we measure the causal contribution of
visual tokens to the model's answer prediction. Given an input sample \(x\),
the vision encoder and multimodal projector produce a visual token sequence
\(V=(v_1,\ldots,v_J)\), where \(J\) denotes the total number of
visual tokens across all input images in \(x\). This sequence is concatenated
with the text tokens \(T\) and fed into the language decoder. We focus on the prefill stage immediately before autoregressive generation and use the logit assigned to the first gold answer token $g_1$ as the information signal, where $g_1$ is the single token representing the gold answer option of each multiple-choice question~\cite{wang2026tokenpruning}, thereby measuring each visual token's contribution to the final answer decision rather than the full reasoning trajectory.

We intervene on the visual sequence at the input-embedding level. Specifically,
masking a visual token means replacing its embedding with a zero vector while
keeping its position in the sequence unchanged. Let \(\widetilde{V}\) denote a
modified visual sequence after masking. We define
\[
\ell(g_1 \mid \widetilde{V}, T)
\]
as the logit assigned to \(g_1\) under the modified visual sequence
\(\widetilde{V}\) and text input \(T\).

We define two reference logits:
\[
\ell_{\mathrm{full}} = \ell(g_1 \mid V, T),
\]
which is the gold-answer logit under the original visual input, and
\[
\ell_{\mathrm{text}} = \ell(g_1 \mid \mathbf{0}_V, T),
\]
which is the logit obtained when all visual token embeddings are masked. The
latter serves as a visual-ablated reference.

\begin{figure}[t]
  \centering
  \includegraphics[width=\linewidth]{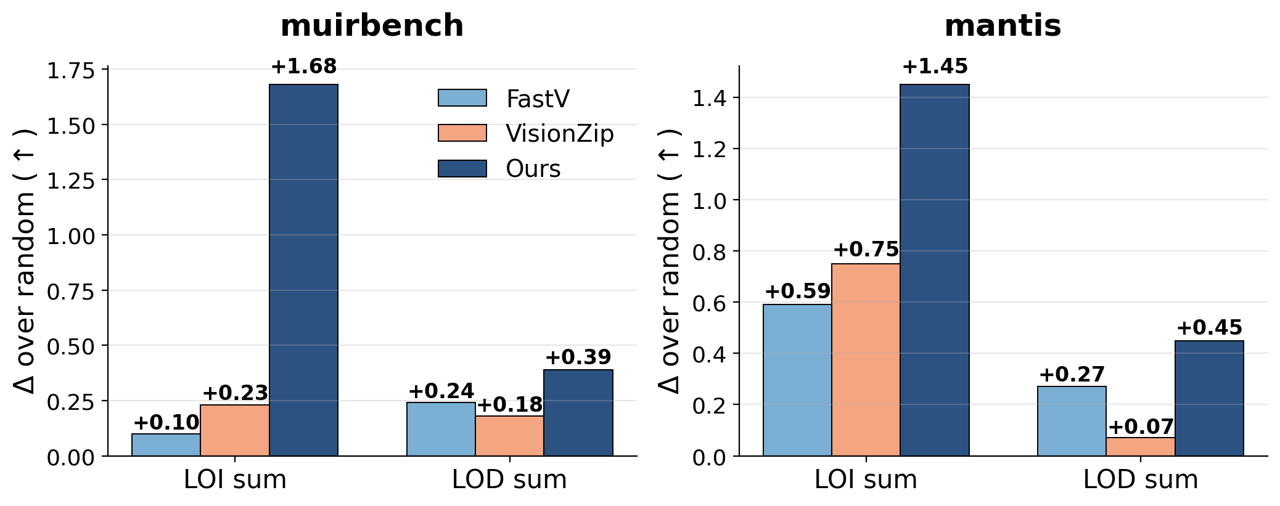}
  \caption{
    \textbf{Method-level information gain over random pruning} when retaining
$4.6\%$ of visual tokens.
    Bars report the per-sample average gains
    $\Delta\mathcal{I}_{\mathrm{LOI}}$ and
    $\Delta\mathcal{I}_{\mathrm{LOD}}$ over uniformly random pruning under the
    same token budget.
    RI-Prune achieves the largest gains across both benchmarks and metrics,
    indicating that it retains more answer-relevant visual tokens than
    attention-based pruning strategies.
  }
  \label{fig:loi_lod_delta}
\end{figure}

\noindent\textbf{Leave-One-In information.}
For each visual token \(v_k\), we construct \(V^{\mathrm{in}}_k\), where all
visual tokens are masked except \(v_k\). The resulting gold-answer logit is
\[
\ell_k^{\mathrm{LOI}} = \ell(g_1 \mid V^{\mathrm{in}}_k, T).
\]
We define the Leave-One-In information score as
\[
I_{\mathrm{LOI}}(v_k)
=
\ell_k^{\mathrm{LOI}} - \ell_{\mathrm{text}}.
\]
A larger LOI score indicates that the token provides stronger standalone
evidence for the correct answer beyond the visual-ablated reference.

\noindent\textbf{Leave-One-Deletion information.}
For each visual token \(v_k\), we construct \(V^{\mathrm{del}}_k\), where only
\(v_k\) is masked and all other visual tokens remain unchanged. The resulting
gold-answer logit is
\[
\ell_k^{\mathrm{LOD}} = \ell(g_1 \mid V^{\mathrm{del}}_k, T).
\]
We define the Leave-One-Deletion information score as
\[
I_{\mathrm{LOD}}(v_k)
=
\ell_{\mathrm{full}} - \ell_k^{\mathrm{LOD}}.
\]
A larger LOD score indicates that removing the token causes a larger decrease
in the gold-answer logit, suggesting a stronger marginal contribution in the
full visual context.

\noindent\textbf{Method-level information score.}
Given a token budget $B$, each pruning method $\mathcal{M}$ selects a token
index set $S_{\mathcal{M}}(x,B) \subseteq \{1,\ldots,J\}$ for each sample
$x$, where $|S_{\mathcal{M}}(x,B)|=B$. We aggregate the per-token scores
over the selected tokens:
\[
\mathcal{I}_{\ast}(\mathcal{M};x,B)
=
\sum_{k \in S_{\mathcal{M}}(x,B)}
I_{\ast}(v_k),
\qquad
\ast \in \{\mathrm{LOI}, \mathrm{LOD}\}.
\]
Here, $\mathcal{I}_{\mathrm{LOI}}$ measures the aggregate standalone evidence of the retained tokens, while $\mathcal{I}_{\mathrm{LOD}}$ measures their aggregate marginal contribution score in the full visual context.

To report gains over random pruning, we compare each method against a uniformly
random subset of the same budget.
Higher gain indicates that the pruning method retains
more answer-relevant visual tokens than random selection under the same budget.

\subsection{Attention Scores Fail to Preserve Informative Visual Tokens.}
Figure~\ref{fig:loi_lod_delta} compares the method-level LOI and LOD gains over ran-
dom pruning under the same token budget. The results reveal a clear qualitative gap between our method and the existing
pruning baselines in terms of how well they preserve answer-relevant visual
evidence. While FastV and VisionZip yield only marginal gains over random
pruning, often staying close to the random baseline, our method consistently
retains visual tokens that carry substantially stronger causal contribution to
the gold answer prediction on both benchmarks. This indicates that the
baselines, which rely on attention-based or feature-redundancy heuristics, tend
to select tokens that are statistically salient inside the model rather than
genuinely informative for the multi-image reasoning task. A more important observation arises from the contrast between the two
measurements. The improvement in LOI is considerably larger than that in LOD,
which suggests that the baselines particularly struggle to identify tokens that
carry standalone answer-relevant evidence. They tend to keep tokens whose
importance is diluted within the redundant visual context. In contrast, our
method preserves tokens that not only contribute marginally in the full context
(LOD) but also independently support the correct prediction (LOI). This
property is critical in the multi-image setting, where each retained token
should ideally encode complementary information rather than reinforce already
redundant visual cues.

Taken together, these results suggest that effective visual token pruning for
multi-image reasoning requires moving beyond intra-image saliency or feature
redundancy, and instead explicitly accounting for the causal contribution of
each token to the answer. Our method aligns with this principle, and the
larger gain over random pruning reflects that the retained tokens are
semantically more faithful to the evidence the model actually relies on when
producing the correct answer.

\section{Method}

\begin{figure*}[t]
    \centering
    \includegraphics[width=\textwidth]{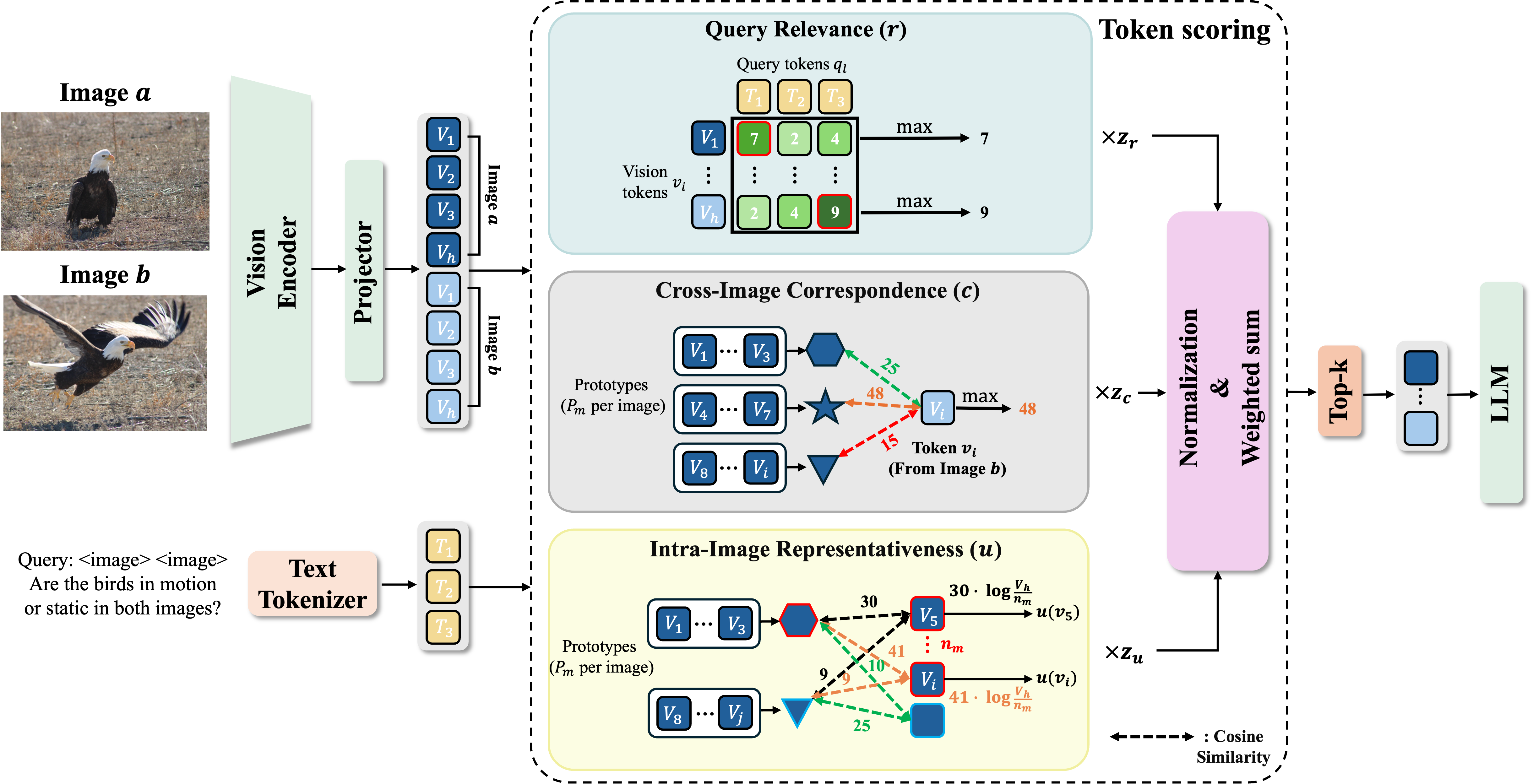}
    \caption{
    Overview of the proposed visual token selection framework.
    Query relevance ($r$), cross-image correspondence ($c$), and intra-image representativeness ($u$) are computed, normalized, and aggregated to select the Top-$K$ vision tokens for the LLM.
    }
    \label{fig:token-selection}
\end{figure*}

\subsection{Overview}

Given \(N\) input images and a text query, the vision encoder extracts image features, which are then mapped into the language model's embedding space by a multimodal projector. This yields a visual token sequence \(V = \{v_1, v_2, \dots, v_J\}\), where \(v_i \in \mathbb{R}^{d}\). The visual tokens are concatenated with the text token embeddings \(T\) and subsequently fed into the language decoder.

RI-Prune operates exactly at this interface: after the projector and before the decoder. Therefore, all downstream decoder computation is performed only on the retained visual tokens. The pruning procedure consists of two steps. First, RI-Prune computes three complementary token-level scores from the projected visual tokens and text embeddings. Second, it selects a subset of visual tokens under a fixed budget \(B\). The selected tokens are forwarded to the language decoder in their original order, while the remaining tokens are discarded.

\subsection{Token Scoring}

RI-Prune assigns each visual token \(v_i\) a final importance score by combining three training-free signals: query relevance, cross-image correspondence, and intra-image representativeness. These signals capture complementary requirements of multi-image reasoning:
task-relevant evidence should be aligned with the question, relational evidence
should be connected to content across images, and redundant local patterns
should be suppressed while preserving representative content within each image.
Rather than relying on a single heuristic signal, RI-Prune combines these three
aspects to estimate token utility from the projected multimodal representations.

Let \(V^{(a)}\) denote the set of visual tokens originating from image \(a\), where \(a \in \{1,\dots,N\}\). All similarity scores are computed using cosine similarity between \(\ell_2\)-normalized embeddings.

\noindent\textbf{Query relevance.}
Let \(Q=\{q_1,q_2,\dots,q_L\}\subseteq T\) denote the embeddings of the textual query tokens. For each visual token \(v_i\), we define its query-relevance score as the maximum cosine similarity to any query-token embedding:
\[
r(v_i) = \max_{1 \le \ell \le L} \cos(v_i, q_\ell).
\]
This score measures direct semantic alignment between visual content and the
task instruction. Since the multimodal projector maps visual representations into the
language-model embedding space, they can be directly compared with
text-token embeddings using cosine similarity, providing a simple
training-free measure of cross-modal alignment. By taking the token-wise maximum, the score preserves fine-grained alignments between visual regions and individual query concepts—such as objects, attributes, or relations.

\noindent\textbf{Cross-image correspondence.}
Multi-image reasoning often requires identifying visual evidence that is
shared, contrasted, or related across images. Query relevance alone does not
explicitly capture such inter-image structure; therefore, we introduce a
cross-image correspondence signal that measures whether a token has
semantically related visual evidence in another image. To compute this signal
efficiently, we summarize each image using a compact set of prototypes.
For each image $a$, we partition its visual tokens $V^{(a)}$
into $P$ contiguous chunks of approximately equal size. We then average the tokens in each
chunk to obtain the prototype set
\(\mathcal{P}^{(a)}=\{p_1^{(a)}, p_2^{(a)}, \dots, p_P^{(a)}\}\), where
\(\mathcal{P}^{(a)}\) denotes the set of prototypes for image \(a\), and
\(p_m^{(a)} \in \mathbb{R}^{d}\) denotes the \(m\)-th prototype vector.

For a token \(v_i \in V^{(a)}\), its cross-image correspondence score is
defined as the maximum cosine similarity to prototypes from the other images:
\[
c(v_i)
=
\max_{\substack{b \neq a \\ 1 \le m \le P}}
\cos(v_i, p_m^{(b)}).
\]
A high value of \(c(v_i)\) indicates that the content represented by \(v_i\) is
semantically consistent with some region in another image, making it a useful
candidate for cross-image comparison or aggregation.

\noindent\textbf{Intra-image representativeness.}
The third signal aims to preserve informative content within each image by emphasizing tokens that are strongly associated with less frequently matched prototypes. This allows RI-Prune to retain representative local evidence while reducing the influence of pervasive patterns that are likely to correspond to repeated background or common context.

For each token \(v_i \in V^{(a)}\), we first assign it to the closest prototype within the same image:
\[
m_i
=
\arg\max_{1 \le m \le P}
\cos(v_i, p_m^{(a)}).
\]
We then compute a non-negative prototype-alignment score:
\[
s_i
=
\max\left(0, \cos(v_i, p_{m_i}^{(a)})\right),
\]
which measures how strongly \(v_i\) aligns with its assigned prototype. Next, we count the number of tokens in image \(a\) assigned to each prototype:
\[
n_m^{(a)}
=
\left|
\{j : v_j \in V^{(a)},\; m_j = m\}
\right|.
\]
The intra-image representativeness score is defined as
\[
u(v_i)
=
s_i \cdot
\log
\frac{|V^{(a)}|}{n_{m_i}^{(a)}}.
\]
The first factor rewards tokens that are confidently associated with a prototype, while the logarithmic factor upweights tokens assigned to less frequent prototypes within the image. Thus, this score favors tokens that represent distinctive yet coherent local patterns. Since \(u(v_i)\) is combined with query relevance and cross-image correspondence rather than used as a standalone criterion, visually distinctive but query-irrelevant outliers are less likely to dominate the final selection.

\noindent\textbf{Final token score.}
The three raw scores have different scales and distributions. We therefore normalize each score within the current sample using z-score normalization. For each scoring function \(x \in \{r,c,u\}\), we define
\[
z_x(v_i)
=
\frac{x(v_i)-\mu_x}{\sigma_x+\epsilon},
\]
where \(\mu_x\) and \(\sigma_x\) are computed over all visual tokens in the current input, and \(\epsilon\) is a small constant for numerical stability. The final importance score of token \(v_i\) is computed as
\[
S(v_i)
=
\lambda_r z_r(v_i)
+
\lambda_c z_c(v_i)
+
\lambda_u z_u(v_i),
\]
where \(\lambda_r\), \(\lambda_c\), and \(\lambda_u\) are non-negative weights controlling the contribution of each signal.

\subsection{Token Selection}

Given a global visual token budget \(B\), RI-Prune selects tokens according to
the final score \(S(\cdot)\). To ensure that every input image contributes
visual evidence, we distribute the budget nearly evenly across the \(N\) images.
Let \(B^{(a)}\) denote the quota assigned to image \(a\), where
\(\sum_{a=1}^{N} B^{(a)} = B\) and the quotas differ by at most one token.

Within each image, we retain the \(B^{(a)}\) tokens with the highest final
importance scores:
\[
R^{(a)}
=
\operatorname{TopK}_{B^{(a)}}
\left(
\{v_i \in V^{(a)}\}; S
\right).
\]
The final retained token set is
\[
R = \bigcup_{a=1}^{N} R^{(a)}.
\]
The retained tokens are then passed to the language decoder in their original
sequence order. This per-image allocation preserves evidence from all input
images, while the scoring function selects the most query-relevant,
cross-image-consistent, and intra-image representative tokens within each image.

\section{Experiments}

\subsection{Experimental Setup}

\noindent\textbf{Baselines and models.}
To verify the generalizability of RI-Prune as a plug-and-play visual token pruning method, we integrate it into representative LVLMs and evaluate it across diverse multi-image reasoning tasks. Specifically, we adopt Qwen2.5-VL-7B-Instruct~\cite{bai2025qwen25vl}, Qwen3-VL-8B-Instruct~\cite{yang2025qwen3} and LLaVA-OneVision-7B~\cite{li2024llavaonevision} as the backbone LVLMs, all of which natively support multi-image inputs. We compare RI-Prune with several plug-and-play, inference-time pruning baselines: FastV~\cite{chen2024image} and SCOPE~\cite{deng2025scope}, which score tokens via decoder attention; VisionZip~\cite{yang2025visionzip}, which selects tokens using encoder-side saliency signals; DivPrune~\cite{alvar2025divprune}, which selects tokens based on visual diversity; and VTW~\cite{lin2025boosting}, which discards all visual tokens at a layer determined by validation performance. 

\noindent\textbf{Benchmarks and evaluation.}
We conduct extensive experiments on four multi-image benchmarks that collectively cover diverse aspects of multi-image reasoning: MUIRBench~\cite{wang2025muirbench}, which spans a wide range of inter-image relations and task types; MIBench~\cite{liu2024mibench}, a large-scale benchmark for multi-image instruction following; MIRB~\cite{zhao2024mirb}, which targets fine-grained perception and reasoning across multiple images; and Mantis-Eval~\cite{jiang2024mantis}, which evaluates multi-image understanding across heterogeneous inter-image tasks. For each benchmark, we report the official task-specific metric. In addition to
accuracy, we evaluate all methods in terms of FLOPs and end-to-end inference
latency, which includes visual encoding, token selection, and decoder computation,
to assess their efficiency under aggressive token reduction.
The visual token reduction ratio is defined as $1-\bar{B}/J$, where $J$ is the original number of visual tokens and $\bar{B}$ denotes the average number of retained tokens across decoder layers. We conduct the main comparison in a high-reduction regime, where differences
between selection strategies are more clearly observed, and further analyze the
robustness of each method across different token budgets. 

\noindent\textbf{Implementation details.}
All pruning experiments are conducted on a single NVIDIA A100 GPU using the HuggingFace Transformers library. RI-Prune is applied once at the projector output, before the first decoder layer, so that all subsequent decoder computation is performed only on the retained tokens. Unless otherwise noted, we set $P=16$ and use $\lambda_r = \lambda_c = \lambda_u = 1$ for all benchmarks and models. For methods that perform progressive or layer-wise reduction, we follow the protocol of prior work and adjust their hyperparameters so that $\bar{B}$ matches our single-shot budget $B$, ensuring that all methods operate under an equivalent token budget. All experiments are conducted with FlashAttention acceleration enabled for both the baselines and RI-Prune to ensure that the reported FLOPs and latency measurements reflect a realistic and consistent inference setting.

\begin{table}[t]
\centering
\caption{
Performance comparison across multi-image benchmarks under different pruning ratios.
Each cell reports accuracy on the first line and relative performance on the second line.
}
\label{tab:main_results}

\fontsize{4.5pt}{4.9pt}\selectfont
\setlength{\tabcolsep}{2.1pt}
\renewcommand{\arraystretch}{0.76}

\def\res#1#2{%
  \shortstack[c]{#1\\[-0.18ex]#2\%}%
}
\def\bres#1#2{%
  \shortstack[c]{\textbf{#1}\\[-0.18ex]\textbf{#2\%}}%
}
\def\conf#1{{\fontsize{2.9pt}{3.1pt}\selectfont\,#1}}
\def\method#1#2{%
  \raisebox{1.25ex}{#1\conf{(#2)}}%
}
\def\methodonly#1{%
  \raisebox{1.25ex}{#1}%
}
\def\methodrule{\specialrule{0.15pt}{0.6pt}{0.6pt}}

\def\grouprule{\specialrule{0.45pt}{0.8pt}{0pt}}
\def\groupspace{\noalign{\vskip 0.8pt}}
\def\groupheader#1{%
\grouprule
\rowcolor{gray!15}
\multicolumn{6}{c}{\rule[-0.35ex]{0pt}{1.7ex}\textit{#1}}\\
\groupspace
}

\resizebox{\columnwidth}{!}{%
\begin{tabular}{@{}>{\centering\arraybackslash}m{1.65cm}ccccc@{}}
\toprule
\textbf{Method} & \textbf{Mantis-Eval} & \textbf{MIBench} & \textbf{MIRB} & \textbf{MUIRBench} & \textbf{Avg.} \\
\midrule
\methodonly{Vanilla} & \res{73.3}{100.0} & \res{54.6}{100.0} & \res{63.6}{100.0} & \res{60.9}{100.0} & \res{63.1}{100.0} \\

\groupheader{Retain 45.3\% Tokens}
\method{FastV}{ECCV'24} & \res{71.9}{98.1} & \res{54.0}{98.9} & \res{56.0}{88.1} & \res{51.5}{84.6} & \res{58.3}{92.5} \\
\methodrule
\method{VisionZip}{CVPR'25} & \res{71.9}{98.1} & \res{52.1}{95.4} & \res{49.5}{77.8} & \res{55.9}{91.8} & \res{57.3}{90.9} \\
\methodrule
\method{VTW}{AAAI'25} & \res{55.8}{76.1} & \res{45.6}{83.5} & \res{42.1}{66.3} & \res{44.4}{72.9} & \res{47.0}{74.4} \\
\methodrule
\method{DivPrune}{CVPR'25} & \res{71.9}{98.1} & \res{54.4}{99.6} & \bres{60.0}{94.3} & \res{59.5}{97.7} & \res{61.4}{97.4} \\
\methodrule
\method{SCOPE}{NeurIPS'25} & \res{72.4}{98.7} & \res{54.4}{99.6} & \res{58.3}{91.7} & \res{57.4}{94.3} & \res{60.6}{96.1} \\
\methodrule
\method{RI-Prune}{Ours} & \bres{72.8}{99.4} & \bres{54.5}{99.8} & \res{58.0}{91.2} & \bres{60.9}{100.0} & \bres{61.5}{97.5} \\

\groupheader{Retain 23.5\% Tokens}
\method{FastV}{ECCV'24} & \res{70.5}{96.2} & \res{52.3}{95.8} & \res{55.1}{86.7} & \res{49.4}{81.1} & \res{56.8}{90.1} \\
\methodrule
\method{VisionZip}{CVPR'25} & \res{68.7}{93.7} & \res{48.5}{88.8} & \res{47.4}{74.5} & \res{54.2}{89.0} & \res{54.7}{86.7} \\
\methodrule
\method{VTW}{AAAI'25} & \res{38.3}{52.2} & \res{26.1}{47.8} & \res{43.3}{68.2} & \res{34.1}{56.0} & \res{35.4}{56.2} \\
\methodrule
\method{DivPrune}{CVPR'25} & \res{71.0}{96.9} & \bres{53.2}{97.4} & \bres{57.5}{90.4} & \res{56.4}{92.6} & \res{59.5}{94.3} \\
\methodrule
\method{SCOPE}{NeurIPS'25} & \res{71.9}{98.1} & \res{52.2}{95.6} & \res{49.4}{77.6} & \res{55.7}{91.5} & \res{57.3}{90.8} \\
\methodrule
\method{RI-Prune}{Ours} & \bres{72.8}{99.4} & \bres{53.2}{97.4} & \bres{57.5}{90.4} & \bres{59.4}{97.5} & \bres{60.7}{96.2} \\

\groupheader{Retain 4.6\% Tokens}
\method{FastV}{ECCV'24} & \res{61.8}{84.3} & \res{33.5}{61.3} & \res{44.0}{69.2} & \res{40.5}{66.5} & \res{44.9}{71.2} \\
\methodrule
\method{VisionZip}{CVPR'25} & \res{59.5}{81.2} & \res{39.0}{71.4} & \res{41.3}{65.0} & \res{47.4}{77.8} & \res{46.8}{74.2} \\
\methodrule
\method{VTW}{AAAI'25} & \res{14.3}{19.5} & \res{22.5}{41.2} & \res{23.0}{36.2} & \res{31.8}{52.2} & \res{22.9}{36.3} \\
\methodrule
\method{DivPrune}{CVPR'25} & \res{62.7}{85.6} & \res{37.5}{68.7} & \res{44.5}{70.1} & \res{49.1}{80.6} & \res{48.4}{76.8} \\
\methodrule
\method{SCOPE}{NeurIPS'25} & \bres{64.5}{88.0} & \bres{39.5}{72.3} & \res{43.2}{68.0} & \res{47.5}{78.0} & \res{48.7}{77.1} \\
\methodrule
\method{RI-Prune}{Ours} & \res{63.6}{86.8} & \res{36.7}{67.2} & \bres{50.3}{79.1} & \bres{52.3}{85.9} & \bres{50.7}{80.4} \\
\bottomrule
\end{tabular}%
}

\vspace{-2mm}
\end{table}


\begin{table}[t]
\centering
\caption{
Comparison with SOTA methods on LLaVA-OneVision-7B under different token budgets.
Each cell reports accuracy on the first line and relative performance on the second line.
Best results are boldfaced, and second-best results are underlined within each token budget.
}
\label{tab:llava_onevision_results}

\fontsize{4.2pt}{4.6pt}\selectfont
\setlength{\tabcolsep}{2.0pt}
\renewcommand{\arraystretch}{0.74}

\def\res#1#2{%
  \shortstack[c]{#1\\[-0.18ex]#2\%}%
}
\def\bres#1#2{%
  \shortstack[c]{\textbf{#1}\\[-0.18ex]\textbf{#2\%}}%
}
\def\ures#1#2{%
  \shortstack[c]{\underline{#1}\\[-0.18ex]\underline{#2\%}}%
}

\def\methodfont#1{{\fontsize{4.9pt}{5.2pt}\selectfont #1}}
\def\conf#1{{\fontsize{2.8pt}{3.0pt}\selectfont\,#1}}
\def\method#1#2{%
  \raisebox{1.25ex}{\methodfont{#1}\conf{(#2)}}%
}
\def\methodonly#1{%
  \raisebox{1.25ex}{\methodfont{#1}}%
}

\def\methodrule{\specialrule{0.15pt}{0.6pt}{0.6pt}}

\def\grouprule{\specialrule{0.45pt}{0.8pt}{0pt}}
\def\groupspace{\noalign{\vskip 0.8pt}}
\def\groupheader#1{%
\grouprule
\rowcolor{gray!15}
\multicolumn{6}{c}{\rule[-0.35ex]{0pt}{1.7ex}\textit{#1}}\\
\groupspace
}

\resizebox{\columnwidth}{!}{%
\begin{tabular}{@{}>{\centering\arraybackslash}m{1.85cm}ccccc@{}}
\toprule
\textbf{Method} & \textbf{Mantis-Eval} & \textbf{MIBench} & \textbf{MIRB} & \textbf{MUIRBench} & \textbf{Avg.} \\
\midrule
\methodonly{Vanilla}
& \res{61.29}{100.0}
& \res{51.37}{100.0}
& \res{52.27}{100.0}
& \res{36.10}{100.0}
& \res{50.3}{100.0} \\

\groupheader{Token Budget 5000}
\method{FastV}{ECCV'24}
& \res{43.9}{71.6}
& \res{42.67}{83.1}
& \res{35.7}{68.3}
& \res{29.6}{82.0}
& \res{38.0}{75.5} \\
\methodrule
\method{VisionZip}{CVPR'25}
& \res{59.0}{96.3}
& \bres{49.87}{97.1}
& \ures{50.0}{95.7}
& \res{34.9}{96.7}
& \ures{48.4}{96.4} \\
\methodrule
\method{VTW}{AAAI'25}
& \res{33.6}{54.8}
& \res{38.00}{74.0}
& \res{39.9}{76.3}
& \res{26.1}{72.3}
& \res{34.4}{68.5} \\
\methodrule
\method{DivPrune}{CVPR'25}
& \ures{60.4}{98.5}
& \ures{48.57}{94.5}
& \res{48.8}{93.4}
& \ures{35.7}{98.9}
& \res{48.4}{96.2} \\
\methodrule
\method{SCOPE}{NeurIPS'25}
& \ures{60.4}{98.5}
& \res{48.17}{93.8}
& \res{49.9}{95.5}
& \res{35.0}{97.0}
& \res{48.4}{96.2} \\
\methodrule
\method{RI-Prune}{Ours}
& \bres{60.7}{99.0}
& \res{46.77}{91.0}
& \bres{51.2}{98.0}
& \bres{36.0}{99.7}
& \bres{48.7}{96.8} \\

\groupheader{Token Budget 3300}
\method{FastV}{ECCV'24}
& \res{42.9}{70.0}
& \res{41.67}{81.1}
& \res{34.7}{66.4}
& \res{26.0}{72.0}
& \res{36.3}{72.3} \\
\methodrule
\method{VisionZip}{CVPR'25}
& \res{56.2}{91.7}
& \bres{48.57}{94.5}
& \res{48.6}{93.0}
& \res{34.5}{95.6}
& \bres{47.0}{93.5} \\
\methodrule
\method{VTW}{AAAI'25}
& \res{28.6}{46.7}
& \res{28.27}{55.0}
& \res{36.8}{70.4}
& \res{22.5}{62.3}
& \res{29.0}{57.8} \\
\methodrule
\method{DivPrune}{CVPR'25}
& \bres{56.7}{92.5}
& \res{46.47}{90.5}
& \res{48.7}{93.2}
& \res{33.9}{93.9}
& \res{46.4}{92.4} \\
\methodrule
\method{SCOPE}{NeurIPS'25}
& \bres{56.7}{92.5}
& \ures{47.67}{92.8}
& \ures{48.8}{93.4}
& \ures{34.7}{96.1}
& \bres{47.0}{93.5} \\
\methodrule
\method{RI-Prune}{Ours}
& \ures{56.4}{92.0}
& \res{46.07}{89.7}
& \bres{49.9}{95.5}
& \bres{34.8}{96.4}
& \ures{46.8}{93.1} \\

\groupheader{Token Budget 1700}
\method{FastV}{ECCV'24}
& \res{39.9}{65.1}
& \res{40.07}{78.0}
& \res{30.1}{57.6}
& \res{25.1}{69.5}
& \res{33.8}{67.2} \\
\methodrule
\method{VisionZip}{CVPR'25}
& \res{53.5}{87.3}
& \bres{47.87}{93.2}
& \res{48.3}{92.4}
& \res{33.0}{91.4}
& \bres{45.7}{90.9} \\
\methodrule
\method{VTW}{AAAI'25}
& \res{21.2}{34.6}
& \res{18.00}{35.0}
& \res{25.0}{47.8}
& \res{15.7}{43.5}
& \res{20.0}{39.8} \\
\methodrule
\method{DivPrune}{CVPR'25}
& \bres{55.3}{90.2}
& \res{46.07}{89.7}
& \res{48.1}{92.0}
& \ures{33.2}{92.0}
& \bres{45.7}{90.9} \\
\methodrule
\method{SCOPE}{NeurIPS'25}
& \res{52.1}{85.0}
& \ures{47.07}{91.6}
& \ures{48.5}{92.8}
& \res{33.1}{91.7}
& \res{45.2}{89.9} \\
\methodrule
\method{RI-Prune}{Ours}
& \ures{54.2}{88.4}
& \res{44.87}{87.3}
& \bres{49.2}{94.1}
& \bres{33.5}{92.8}
& \ures{45.4}{90.4} \\
\bottomrule
\end{tabular}%
}

\vspace{-2mm}
\end{table}


\begin{table}[t]
\centering
\caption{
Comparison with SOTA methods on Qwen3-VL under different retained token ratios.
Each cell reports accuracy on the first line and relative performance on the second line.
Best results are boldfaced, and second-best results are underlined within each retained token ratio.
}
\label{tab:qwen3_vl_results}

\fontsize{5.0pt}{5.4pt}\selectfont
\setlength{\tabcolsep}{2.6pt}
\renewcommand{\arraystretch}{0.82}

\def\res#1#2{%
  \shortstack[c]{#1\\[-0.18ex]#2\%}%
}
\def\bres#1#2{%
  \shortstack[c]{\textbf{#1}\\[-0.18ex]\textbf{#2\%}}%
}
\def\ures#1#2{%
  \shortstack[c]{\underline{#1}\\[-0.18ex]\underline{#2\%}}%
}

\def\methodfont#1{{\fontsize{5.4pt}{5.8pt}\selectfont #1}}
\def\conf#1{{\fontsize{3.1pt}{3.3pt}\selectfont\,#1}}
\def\method#1#2{%
  \raisebox{1.25ex}{\methodfont{#1}\conf{(#2)}}%
}
\def\methodonly#1{%
  \raisebox{1.25ex}{\methodfont{#1}}%
}

\def\methodrule{\specialrule{0.15pt}{0.7pt}{0.7pt}}

\def\grouprule{\specialrule{0.45pt}{0.9pt}{0pt}}
\def\groupspace{\noalign{\vskip 1.0pt}}
\def\groupheader#1{%
\grouprule
\rowcolor{gray!15}
\multicolumn{6}{c}{\rule[-0.35ex]{0pt}{2.0ex}\textit{#1}}\\
\groupspace
}

\resizebox{\columnwidth}{!}{%
\begin{tabular}{@{}>{\centering\arraybackslash}m{2.05cm}ccccc@{}}
\toprule
\textbf{Method} & \textbf{Mantis-Eval} & \textbf{MIBench} & \textbf{MIRB} & \textbf{MUIRBench} & \textbf{Avg.} \\
\midrule

\methodonly{Vanilla}
& \res{78.7}{100.0}
& \res{60.85}{100.0}
& \res{68.75}{100.0}
& \res{64.4}{100.0}
& \res{68.2}{100.0} \\

\groupheader{Retain 23.5\% Tokens}
\method{FastV}{ECCV'24}
& \res{69.4}{88.2}
& \res{55.1}{90.6}
& \res{59.3}{86.3}
& \res{60.3}{93.6}
& \res{61.0}{89.5} \\
\methodrule
\method{VisionZip}{CVPR'25}
& \res{74.6}{94.8}
& \res{53.8}{88.4}
& \res{62.4}{90.8}
& \res{60.9}{94.6}
& \res{62.9}{92.3} \\
\methodrule
\method{VTW}{AAAI'25}
& \res{63.7}{80.9}
& \res{48.0}{78.9}
& \res{40.1}{58.3}
& \res{58.5}{90.8}
& \res{52.6}{77.1} \\
\methodrule
\method{DivPrune}{CVPR'25}
& \res{64.7}{82.2}
& \res{51.2}{84.1}
& \res{60.1}{87.4}
& \res{61.0}{94.7}
& \res{59.3}{86.9} \\
\methodrule
\method{SCOPE}{NeurIPS'25}
& \ures{75.1}{95.4}
& \bres{58.7}{96.5}
& \ures{63.0}{91.6}
& \ures{62.7}{97.4}
& \ures{64.9}{95.2} \\
\methodrule
\method{RI-Prune}{Ours}
& \bres{75.5}{95.9}
& \ures{57.2}{94.0}
& \bres{66.0}{96.0}
& \bres{64.1}{99.5}
& \bres{65.7}{96.4} \\

\groupheader{Retain 4.6\% Tokens}
\method{FastV}{ECCV'24}
& \res{60.4}{76.7}
& \res{45.5}{74.8}
& \res{52.1}{75.8}
& \res{49.6}{77.0}
& \res{51.9}{76.1} \\
\methodrule
\method{VisionZip}{CVPR'25}
& \res{68.2}{86.7}
& \ures{51.2}{84.1}
& \res{52.9}{76.9}
& \res{57.3}{89.0}
& \res{57.4}{84.2} \\
\methodrule
\method{VTW}{AAAI'25}
& \res{54.2}{68.9}
& \res{43.5}{71.5}
& \ures{62.1}{90.3}
& \res{42.3}{65.7}
& \res{50.5}{74.1} \\
\methodrule
\method{DivPrune}{CVPR'25}
& \res{54.2}{68.9}
& \res{47.6}{78.2}
& \res{52.2}{75.9}
& \ures{59.9}{93.0}
& \res{53.5}{78.4} \\
\methodrule
\method{SCOPE}{NeurIPS'25}
& \bres{74.7}{94.9}
& \res{47.5}{78.1}
& \res{52.5}{76.4}
& \res{58.3}{90.5}
& \ures{58.3}{85.4} \\
\methodrule
\method{RI-Prune}{Ours}
& \ures{73.2}{93.0}
& \bres{53.2}{87.4}
& \bres{65.8}{95.7}
& \bres{61.6}{95.7}
& \bres{63.5}{93.1} \\
\bottomrule
\end{tabular}%
}

\vspace{-2mm}
\end{table}

\subsection{Experimental Results}
\noindent\textbf{Results on Qwen2.5-VL.}
We evaluate all pruning methods on Qwen2.5-VL across four multi-image benchmarks, as shown in Table~\ref{tab:main_results}. The vanilla model achieves an average score of $63.1$, which serves as the reference for evaluating performance retention. RI-Prune maintains strong performance across all three retained token ratios. At $45.3\%$ retained tokens, it achieves an average score of $61.5$, corresponding to $97.5\%$ of the vanilla performance, and slightly outperforms the strongest baseline, DivPrune. It also performs best on Mantis-Eval, MIBench, and MUIRBench. When the retained ratio is reduced to $23.5\%$, RI-Prune obtains an average score of $60.7$ ($96.2\%$ of vanilla), improving over DivPrune by $1.2$ points and achieving the best or tied-best result across all four benchmarks. This suggests that its advantage is maintained as the available visual token budget becomes smaller. Under the most aggressive setting, with only $4.6\%$ of visual tokens retained, RI-Prune achieves an average score of $50.7$ and preserves $80.4\%$ of the vanilla performance, outperforming SCOPE and DivPrune. In particular, its stronger results on MIRB and MUIRBench indicate that the selected tokens remain effective for fine-grained and cross-image reasoning even under severe token reduction.

\noindent\textbf{Results on LLaVA-OneVision.}
To evaluate whether RI-Prune generalizes to a different multi-image LVLM architecture, we further conduct experiments on LLaVA-OneVision-7B under token budgets of $5000$, $3300$, and $1700$, as reported in Table~\ref{tab:llava_onevision_results}. With a budget of $5000$, RI-Prune achieves the highest average score of $48.7$, preserving $96.8\%$ of the vanilla performance, while obtaining the best results on Mantis-Eval, MIRB, and MUIRBench. When the budget is reduced to $3300$, it remains competitive with an average score of $46.8$ ($93.1\%$ of vanilla), ranking second overall while still performing best on MIRB and MUIRBench. At the most constrained budget of $1700$, RI-Prune achieves an average score of $45.4$ and retains $90.4\%$ of the vanilla performance, again showing the strongest results on MIRB and MUIRBench. Overall, these results indicate that RI-Prune transfers effectively to LLaVA-OneVision and remains robust as the visual token budget is progressively reduced.

\noindent\textbf{Results on Qwen3-VL.}
We further evaluate RI-Prune on Qwen3-VL to examine whether its effectiveness extends to a more recent and stronger LVLM backbone. Experiments are conducted at retained token ratios of $23.5\%$ and $4.6\%$, with results reported in Table~\ref{tab:qwen3_vl_results}. We focus on these higher-reduction settings to evaluate generalization under more challenging visual-token constraints. At $23.5\%$, RI-Prune achieves the highest average score of $65.7$, preserving $96.4\%$ of the vanilla performance and improving over SCOPE, the strongest baseline, by $0.8$ points. It also obtains the best results on Mantis-Eval, MIRB, and MUIRBench. The difference becomes larger under the more aggressive $4.6\%$ setting, where RI-Prune reaches an average score of $63.5$ and retains $93.1\%$ of the vanilla performance. This is $5.2$ points higher than SCOPE, while RI-Prune also performs best on MIBench, MIRB, and MUIRBench. These results suggest that the proposed token selection strategy remains effective on a stronger backbone, particularly when the visual token budget is highly constrained.

\begin{table}[t]
\centering
\caption{
Ablation study of each component and its corresponding weighting coefficient in RI-Prune.
Best result is boldfaced.
}
\label{tab:ablation}

\fontsize{6.0pt}{6.5pt}\selectfont
\setlength{\tabcolsep}{4.0pt}
\renewcommand{\arraystretch}{1.05}

\def\ck{\checkmark}
\def\abest#1{\textbf{#1}}

\resizebox{\columnwidth}{!}{%
\begin{tabular}{@{}>{\centering\arraybackslash}m{1.55cm}ccc ccc cc@{}}
\toprule
\textbf{Variant} 
& $\boldsymbol{\lambda_r}$ & $\boldsymbol{\lambda_c}$ & $\boldsymbol{\lambda_u}$
& $\boldsymbol{r}$ & $\boldsymbol{c}$ & $\boldsymbol{u}$
& \textbf{Acc.} & $\boldsymbol{\Delta}$ \textbf{vs Full} \\
\midrule

Random 
& 0 & 0 & 0 
&  &  &  
& 40.1 & $-10.6$ \\

$r$ only 
& 1 & 0 & 0 
& \ck &  &  
& 44.6 & $-6.1$ \\

$c$ only 
& 0 & 1 & 0 
&  & \ck &  
& 43.1 & $-7.6$ \\

$u$ only 
& 0 & 0 & 1 
&  &  & \ck 
& 42.4 & $-8.3$ \\

$r+c$ 
& 1 & 1 & 0 
& \ck & \ck &  
& 48.6 & $-2.1$ \\

$r+u$ 
& 1 & 0 & 1 
& \ck &  & \ck 
& 41.8 & $-8.9$ \\

$c+u$ 
& 0 & 1 & 1 
&  & \ck & \ck 
& 45.5 & $-5.2$ \\

$\lambda_r=0.5$
& 0.5 & 1 & 1
& \ck & \ck & \ck
& 49.5 & $-1.2$ \\

$\lambda_c=0.5$
& 1 & 0.5 & 1
& \ck & \ck & \ck
& 49.8 & $-0.9$ \\

$\lambda_u=0.5$
& 1 & 1 & 0.5
& \ck & \ck & \ck
& 50.2 & $-0.5$ \\

\midrule
\abest{Full $(r+c+u)$} 
& 1 & 1 & 1 
& \ck & \ck & \ck 
& \abest{50.7} & -- \\

\bottomrule
\end{tabular}%
}

\vspace{-2mm}
\end{table}

\subsection{Ablation analysis}
\noindent\textbf{Ablation Studies.}
Table~\ref{tab:ablation} analyzes the contribution and weighting of each RI-Prune component on Qwen2.5-VL at the 4.6\% retained-token setting, reporting the average accuracy across the four multi-image benchmarks. The random baseline achieves $40.1$, while all single-component variants improve over it, with $r$-only performing best at $44.6$, followed by $c$-only and $u$-only. Among partial combinations, $r+c$ achieves the strongest result ($48.6$), reducing the gap to the full model to $2.1$ points, while $c+u$ also improves over its individual components. We further vary each weighting coefficient from $1$ to $0.5$ while keeping the others fixed. The resulting accuracies remain high ($49.5$--$50.2$) but consistently below the equal-weight setting. The full configuration with $\lambda_r{=}\lambda_c{=}\lambda_u{=}1$ achieves the best accuracy of $50.7$, outperforming the strongest partial variant by $2.1$ points and random selection by $10.6$ points. These results indicate that the three signals are complementary and that equal weighting provides a simple and robust configuration.

\begin{table}[t]
\centering
\caption{
Efficiency and accuracy comparison under different retained token ratios.
Results are reported on Qwen2.5-VL.
}
\label{tab:efficiency_results}

\scriptsize
\setlength{\tabcolsep}{3.0pt}
\renewcommand{\arraystretch}{0.92}

\newcommand{\best}[1]{\textbf{#1}}
\newcommand{\second}[1]{\underline{#1}}

\resizebox{\columnwidth}{!}{%
\begin{tabular}{lcccc}
\toprule
\textbf{Method} & \textbf{Retain Ratio} & \textbf{TFLOPs} & \textbf{Latency} & \textbf{Avg. Acc.} \\
\midrule
Qwen2.5-VL & 100.0\% & 57.8 & 8.135s & 63.1 \\
\midrule

FastV & 4.6\% & 11.4 & 7.728s & 44.9 \\
DivPrune & 4.6\% & 8.8 & 8.074s & 48.4 \\
SCOPE & 4.6\% & 8.9 & 7.953s & \second{48.7} \\
VisionZip & 4.6\% & 8.4 & 5.217s & 46.8 \\
VTW & 4.6\% & 8.5 & 5.125s & 22.9 \\
\textbf{RI-Prune} & 4.6\% & \best{7.9} & \best{5.124s} & \best{50.7} \\
\midrule

FastV & 23.5\% & 13.4 & 8.128s & 56.8 \\
DivPrune & 23.5\% & 10.79 & 8.774s & \second{59.5} \\
SCOPE & 23.5\% & 10.79 & 8.153s & 57.3 \\
VisionZip & 23.5\% & \best{10.35} & 5.557s & 54.7 \\
VTW & 23.5\% & 10.64 & 5.636s & 35.4 \\
\textbf{RI-Prune} & 23.5\% & \best{10.35} & \best{5.554s} & \best{60.7} \\
\bottomrule
\end{tabular}%
}

\vspace{-2mm}
\end{table}


\noindent\textbf{Efficiency Analysis.}
Table~\ref{tab:efficiency_results} compares the efficiency and accuracy of different pruning methods on Qwen2.5-VL under two retained token ratios. The full model achieves an average accuracy of $63.1$ with $57.8$ TFLOPs and $8.135$s latency. When retaining $23.5\%$ of visual tokens, RI-Prune reduces computation to $10.35$ TFLOPs, corresponding to a $5.6\times$ reduction over the full model, while maintaining an average accuracy of $60.7$ and preserving $96.2\%$ of the full-model performance. It outperforms DivPrune by $1.2$ accuracy points while requiring lower inference latency and matching the lowest TFLOPs among the compared methods. This indicates that substantial visual token reduction can be achieved without a large degradation in task performance. Under the more aggressive $4.6\%$ retention ratio, RI-Prune achieves the best average accuracy of $50.7$ while using $7.9$ TFLOPs and $5.124$s latency. This corresponds to a $7.3\times$ reduction in TFLOPs and a $1.59\times$ speedup over the full model. The relative performance advantage becomes clearer in this high-compression setting, suggesting that RI-Prune is particularly effective when only a small visual token budget is available. Overall, these results show that RI-Prune provides a favorable balance between computational efficiency and task performance under aggressive token reduction.

\begin{table}[t]
\centering
\caption{
Comparison of token selection modes on Qwen2.5-VL at the $4.6\%$ retained-token
setting. Selection overhead is measured in seconds, excluding decoder forward
computation.
}
\label{tab:selection_mode}

\scriptsize
\setlength{\tabcolsep}{5.0pt}
\renewcommand{\arraystretch}{0.95}

\resizebox{\columnwidth}{!}{%
\begin{tabular}{lccc}
\toprule
\textbf{Selection Mode} & \textbf{Accuracy} & \textbf{Selection Overhead (s)} & \textbf{TFLOPs} \\
\midrule
Global Top-$K$ 
& 48.8 & \textbf{0.909} & 7.9 \\

Global $+$ min quota 
& 50.3 & 0.922 & 7.9 \\

Per-image Top-$K$ 
& \textbf{50.7} & 0.920 & 7.9 \\
\bottomrule
\end{tabular}%
}

\vspace{-2mm}
\end{table}

\noindent\textbf{Selection Mode Analysis.}
Table~\ref{tab:selection_mode} compares different token selection modes under
the same token budget. Since all modes use the same number of retained tokens
and have identical TFLOPs with only minor selection overhead latency differences, the performance
difference mainly comes from how the token budget is distributed across images.
The global Top-$K$ strategy achieves the lowest latency but the lowest accuracy,
suggesting that selecting tokens from a flat global pool can cause visually
dominant images to monopolize the budget while leaving other images
underrepresented. This is undesirable for multi-image reasoning, where the
answer often depends on evidence distributed across multiple images. Introducing a minimum per-image quota alleviates this issue by ensuring basic
coverage for each image, leading to a clear improvement over pure global
selection. However, because the remaining budget is still allocated globally,
the method can still be biased toward a subset of images. In contrast, the
per-image Top-$K$ strategy performs best while maintaining comparable latency
and the same TFLOPs. By preserving the most informative tokens within each image,
it better respects the multi-image structure of the input and retains
complementary cross-image evidence. These results suggest that balanced
per-image allocation is a more suitable inductive bias for multi-image LVLM
pruning than treating all visual tokens as a single global candidate pool.
    
\begin{table}[t]
\centering
\caption{
Performance comparison according to the number of prototypes $P$ on Qwen2.5-VL at a 4.6\% retained token budget.
}
\label{tab:hyper_k}

\fontsize{6.2pt}{6.7pt}\selectfont
\setlength{\tabcolsep}{5.0pt}
\renewcommand{\arraystretch}{1.05}

\resizebox{\columnwidth}{!}{%
\begin{tabular}{@{}lcccccc@{}}
\toprule
\textbf{$P$} 
& \textbf{2} & \textbf{4} & \textbf{8} & \textbf{16} & \textbf{64} & \textbf{128} \\
\midrule
\textbf{Accuracy}
& 45.6 & 46.9 & 47.7 & 50.7 & 51.0 & 48.7 \\

\textbf{Compute (s)}
& 0.0034 & 0.0035 & 0.0038 & 0.0044 & 0.0082 & 0.0116 \\
\bottomrule
\end{tabular}%
}

\vspace{-2mm}
\end{table}

\noindent\textbf{Effect of the number of prototypes $\boldsymbol{P}$.}
Table~\ref{tab:hyper_k} reports the accuracy and per-sample compute time as we
vary the number of prototypes $P$. Accuracy improves consistently as $P$
grows from $2$ to $16$, indicating that a sufficient
number of prototypes is required to cover the diverse visual semantics that
appear across multiple images. The peak accuracy is reached at $P{=}64$, but the improvement over $P{=}16$ is marginal while the
compute cost nearly doubles. Further increasing $P$ to $128$ degrades accuracy to $48.7$, which
suggests that an excessive number of prototypes fragments semantically coherent
regions and weakens the representativeness of each prototype. Considering the
trade-off between accuracy and computational overhead, we adopt $P{=}16$ as
the default setting, which achieves competitive accuracy with the lowest
compute among the top-performing configurations.

\begin{table}[t]
\centering
\caption{
Accuracy comparison on video benchmarks.
Acc. denotes accuracy (\%). Among pruning methods, the best results are boldfaced
and the second-best results are underlined.
}
\label{tab:video_benchmark_results}

\fontsize{7.4pt}{8.0pt}\selectfont
\setlength{\tabcolsep}{2.4pt}
\renewcommand{\arraystretch}{1.15}

\def\best#1{\textbf{#1}}
\def\second#1{\underline{#1}}

\resizebox{\columnwidth}{!}{%
\begin{tabular}{@{}lccccccc@{}}
\toprule
\textbf{Benchmark}
& \textbf{Vanilla}
& \textbf{FastV}
& \textbf{DivPrune}
& \textbf{SCOPE}
& \textbf{VisionZip}
& \textbf{VTW}
& \textbf{RI-Prune} \\
\midrule
Video-MME
& 63.0
& 49.3
& \second{52.2}
& 47.8
& 44.9
& 23.2
& \best{52.9} \\

TempCompass
& 67.0
& 59.8
& \best{63.0}
& 59.0
& 55.0
& 33.0
& \second{61.0} \\
\bottomrule
\end{tabular}%
}

\vspace{-2mm}
\end{table}

\begin{figure*}[t]
  \centering
  \includegraphics[width=\textwidth]{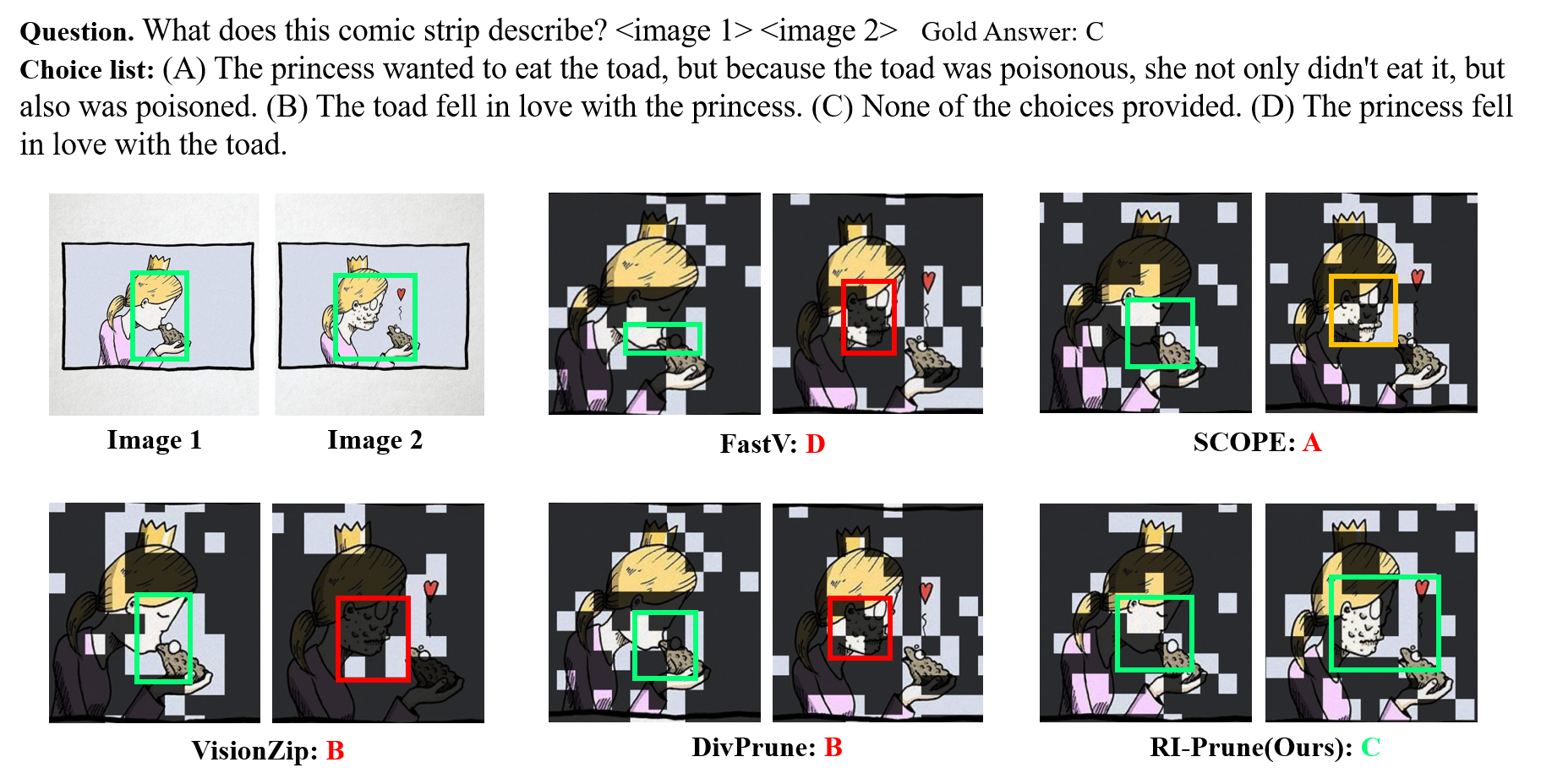}
  \caption{
      \textbf{Qualitative visualization of visual token pruning on MUIRBench~\cite{wang2025muirbench}.}
      Bright regions indicate retained visual tokens. Colored boxes denote
      question-relevant regions: green indicates successful preservation, red
      indicates pruning of critical information, and orange indicates ambiguous
      preservation that ultimately leads to an incorrect prediction.
      RI-Prune preserves more relevant visual evidence under aggressive token reduction.
  }
  \label{fig:pruning_visualization}
\end{figure*}

\noindent\textbf{Results on Video Benchmarks.}
To examine whether RI-Prune generalizes beyond static multi-image inputs, we
evaluate all pruning methods on Video-MME~\cite{fu2025videomme} and TempCompass~\cite{liu2024tempcompass}, which require
reasoning over temporally ordered frames and are structurally analogous to the
multi-image regime. All pruners operate under the same matched token budget as
the $23.5\%$ retained setting. As shown in
Table~\ref{tab:video_benchmark_results}, RI-Prune achieves the best Video-MME
accuracy, surpassing DivPrune and clearly outperforming
attention-based pruners such as FastV, SCOPE, and VisionZip. On TempCompass, RI-Prune ranks second with $61.0$, trailing DivPrune but still outperforming the attention-driven baselines. In contrast,
the layer-truncation baseline VTW performs poorly on both benchmarks, achieving
only $23.2$ on Video-MME and $33.0$ on TempCompass. These results suggest that RI-Prune can extend to video inputs without
modification. Since video frames can be viewed as temporally ordered visual
inputs, the query-relevance signal helps prioritize frame tokens that are more
aligned with the question, while the cross-image-correspondence signal can
capture recurring or related visual content across frames. This may explain the
strong performance on Video-MME. On TempCompass, where fine-grained temporal
attribute discrimination can require broader frame coverage, DivPrune performs
slightly better, but RI-Prune remains competitive. Overall, the consistent gap
between RI-Prune and attention-based pruners suggests that attention-free,
relation-aware token selection is also beneficial for video-style multi-visual
inputs.

\noindent\textbf{Qualitative visualization.}
Figure~\ref{fig:pruning_visualization} visualizes pruning results on MUIRBench
with only $128$ retained visual tokens. While most baselines preserve the
salient evidence in the first image, they often miss the important evidence in
the second image, where the poisoned state of the princess is required for
reasoning. This suggests that conventional pruning methods tend to focus on
locally salient or visually dominant regions, but may fail to preserve evidence
that is less salient yet crucial for cross-image comparison. In contrast, RI-Prune retains relevant regions in both images. This is because
its scoring combines query relevance with cross-image correspondence, allowing
tokens related to the question and shared reasoning context to be preserved
across images. The visualization supports our quantitative results, showing that
RI-Prune better maintains distributed visual evidence under aggressive pruning.

\section{Conclusion}
In this paper, we revisit visual token pruning for multi-image vision-language models and show that attention-based saliency transfers poorly from single-image to multi-image reasoning. Analyses across multiple LVLMs reveal positional bias in query-driven attention and weak query alignment of CLS-based saliency, while causal probes show that attention-based pruners retain only marginally more informative tokens than random selection. Motivated by these findings, we propose RI-Prune, a training-free and attention-free method operating in the projector embedding space. RI-Prune combines query relevance, cross-image correspondence, and intra-image representativeness with per-image Top-$K$ selection. Experiments across multiple benchmarks and backbones demonstrate that RI-Prune consistently preserves task performance under aggressive token reduction, while ablations validate the complementary scoring signals and the effectiveness of per-image allocation.

\noindent\textbf{Limitations and Future Work.}
RI-Prune has several limitations. First, its near-uniform per-image budgets may be suboptimal when relevant evidence is concentrated in only a few images, motivating adaptive content-aware allocation. Second, it operates only at the projector--decoder interface and does not exploit layer-wise redundancy within the decoder. Third, its cross-image correspondence signal may become costly for long interleaved video or document inputs. Finally, our evaluation is limited to 7--8B models and multiple-choice multi-image benchmarks; extending RI-Prune to larger models, long-video understanding, and generation-heavy tasks remains future work.

\begin{acks}
This work was supported by the Institute of Information \& Communications Technology Planning \& Evaluation (IITP) grant funded by the Korea government (MSIT) [RS-2021-II211341, Artificial Intelligence Graduate School Program (Chung-Ang University)] and by the National Research Foundation of Korea (NRF) grant funded by the Korea government (MSIT) (RS-2025-00556246).
\end{acks}

\section*{GenAI Usage Disclosure}
The authors used generative AI tools, including ChatGPT, to assist with
language polishing, grammar correction, wording refinement, and LaTeX formatting
during the preparation of this manuscript. These tools were also used to help
revise captions, tables, and result descriptions based on author-provided
experimental results.

Generative AI tools were not used to generate benchmark data, modify evaluation
datasets, fabricate experimental results, or make scientific conclusions. They
were also not used as an autonomous agent for conducting experiments. All
GenAI-assisted content was carefully reviewed, verified, and edited by the
authors. The authors take full responsibility for the accuracy, integrity, and
originality of the submitted work.


\bibliographystyle{ACM-Reference-Format}
\bibliography{main}

\appendix

\end{document}